\documentclass{article}
\usepackage{ijcai22}

\usepackage{xspace,mfirstuc,tabulary}

\usepackage[utf8]{inputenc} 
\usepackage[T1]{fontenc}    
\usepackage{hyperref}       
\usepackage{url}            
\usepackage{booktabs}       
\usepackage{amsfonts}       
\usepackage{nicefrac}       
\usepackage{microtype}      
\usepackage{xcolor}         
\usepackage{url}            
\usepackage{booktabs}       
\usepackage{amsfonts}       
\usepackage{nicefrac}       
\usepackage{microtype}      
\usepackage{subcaption}
\usepackage{amsmath,amsfonts,amssymb}
\usepackage{breqn}
\usepackage{tabularx}
\usepackage{multirow}
\usepackage{graphicx}
\usepackage{color}
\usepackage{tcolorbox}
\usepackage{CJK}
\usepackage{adjustbox}
\usepackage{xcolor}
\usepackage{colortbl}
\usepackage{multicol}
\usepackage{vwcol} 
\newsavebox{\algleft}
\newsavebox{\algright}
\usepackage{bm}
\usepackage{booktabs}
\usepackage{ctable} 
\usepackage{amsmath,stackengine}
\usepackage[linesnumbered,ruled,vlined]{algorithm2e}
\usepackage{times}
\usepackage{latexsym}
\usepackage{bbm}
\usepackage[utf8]{inputenc} 
\usepackage[T1]{fontenc}    
\usepackage{hyperref}       
\usepackage{url}            
\usepackage{booktabs}       
\usepackage{amsfonts}       
\usepackage{nicefrac}       
\usepackage{microtype}      
\usepackage{xcolor}         

\title{Continual Learning of Natural Language Processing Tasks: A Survey} 

\author{
Zixuan Ke ~~~~~~~~~~~~~ Bing Liu\\ 
 \affiliations
Department of Computer Science, University of Illinois at Chicago\\
\emails
\{zke4, liub\}@uic.edu\\  
}
\begin{document}

\maketitle


\begin{abstract}
Continual learning (CL) is a learning paradigm that emulates the human capability of learning and accumulating knowledge continually without \textit{forgetting} the previously learned knowledge and also \textit{transferring} the learned knowledge to help learn new tasks better. 
This survey presents a comprehensive review and analysis of the recent progress of CL in NLP, which has significant differences from CL in computer vision and machine learning. It covers (1) all CL settings with a taxonomy of existing techniques; (2) \textit{catastrophic forgetting} (CF) \textit{prevention}, (3) \textit{knowledge transfer} (KT), which is particularly important for NLP tasks; and (4) some theory and the hidden challenge of \textit{inter-task class separation} (ICS). (1), (3) and (4) have not been included in the existing survey. Finally, a list of future directions is discussed.\footnote{Preprint. Work in Progress} 

\end{abstract}

\section{Introduction}
\label{sec.intro}


The goal of \textit{continual learning} (CL) is to learn a sequence of tasks incrementally in a neural network \cite{chen2018lifelong}.\footnote{{\color{black}We refer readers to \cite{chen2018lifelong} for the difference between CL and other machine learning paradigms.}} CL is important in many ways. For example, an LM needs to be updated due to the changing environment, e.g., virus mutations and language evolution. A dialogue system needs to learn new skills. 
A sentiment analysis system needs to constantly learn to classify opinions about new products. Since re-training a model from scratch is usually very expensive and time consuming, incrementally updating it with the latest data and emerging domains/tasks 
is critical. 

CL has been researched in machine learning (ML), computer vision (CV) and natural language processing (NLP). Since it does incremental learning of tasks, 
it assumes that once a task $t$ is learned, its training data $D_t$ (at least a majority of it) is no longer accessible. In NLP, a \textit{task} can be an \textit{end-task} 
(e.g., text classification, summarization and information extraction), which is usually supervised~\cite{ke2021achieving,DBLP:conf/aaai/MonaikulCFR21,DBLP:conf/iclr/QinJ22}, or a \textit{domain} corpus 
used to further pre-train\footnote{It also called \textit{domain-adaptive pre-training} or \textit{pre-finetuning}.
}
an language model (LM), which is usually unsupervised~\cite{gururangan2021demix,ke2022cpt,scialom2022continual}. The task types in NLP can be quite different from those in CV due to their different applications. 

There are two general desiderata for a CL system: (1) It should not suffer from \textbf{\textit{catastrophic forgetting}} (CF), which refers to the phenomenon when learning a new task, the model needs to modify the network parameters learned for previous tasks, which may cause the previous tasks to degrade~\cite{mccloskey1989catastrophic}. (2) It should encourage \textbf{\textit{knowledge transfer}} (KT) across tasks, including (a) \textit{forward transfer}, leveraging the knowledge learned from old tasks to help learn the new task, and (b) \textit{backward transfer}, improving the performance of some previous tasks in learning a similar new task. There are also some specific challenges in some CL settings (Sec.
\ref{sec.setting}). For example, one challenge that is largely ignored by the existing research is  \textbf{\textit{inter-task class separation}} (ICS) for the \textit{class-incremental learning} setting, which has just been raised and dealt with~\cite{kim2022theoretical}.

\textbf{Compared to existing surveys.} This survey makes several contributions: (1) it is the first to cover all CL settings and state-of-the-art approaches of a wide range of NLP tasks.
To our knowledge, 
\cite{biesialska2020continual} is the only survey on NLP-based CL. However, as it was published 2 years ago, it has a very small coverage. It only touches the CF problem and has no mention of KT, which is extremely important for NLP as many tasks in a NLP application are similar. It has no mention of different CL settings, which are critical because different settings are for different applications and use different techniques. It also misses many approaches specifically for NLP (e.g., instruction-based methods and KT methods). Since the ICS problem is new, it is not covered previously either. Other CL surveys all focus on CV tasks,\footnote{\cite{DBLP:journals/corr/abs-1909-08383} is only about TIL, \cite{masana2020class,belouadah2021comprehensive} are only about CIL. \cite{mai2022online} is only about online CL. Some cover more settings~\cite{parisi2019continual,hadsell2020embracing,ven2019three,qu2021recent}. \cite{10.1162/neco_a_01433} focuses on biological inspirations and~\cite{lesort2020continual} focuses on autonomous systems.} without considering the contents specific to NLP (e.g., domain-adaptive pre-training and KT.). (2) This survey analyzes and extends the existing CL settings to make them more general and up-to-date.

\section{Settings and Learning Modes of CL}
\label{sec.setting}


Different CL settings and learning modes lead to different specific challenges and approaches.  

\subsection{CL Settings}
\label{sec.setttings}

CL has several settings \cite{ven2019three}.  
For all settings, it learns a sequence of tasks $1, 2, ..., T$, where $T$ is the last task learned. Let $\mathcal{T}=\{1...T\}$. Each task $t \in \mathcal{T}$ has a training set $D_t=\{(x^{(t)}_i),y^{(t)}_i)\}_{i=1}^{N_t}$, where $x^{(t)}_i$ is a training sample, $y^{(t)}_i$ ($\in \mathcal{Y}_t$, the label space of task $t$) is its class label and $N_t$ is the number of samples. Let $\mathcal{X}$ be the input space of all tasks and 
$p(\mathcal{X}_t)$ be the distribution of the input data of task $t$. For any two tasks $t$ and $t'$, we have $p(\mathcal{X}_t) \neq p(\mathcal{X}_{t'}), \forall~t \ne t'$. 


\textbf{(1) Task-incremental learning (TIL).} In TIL, task-identifiers (task-IDs) are available in both training and testing, and classes in the tasks may or may not be disjoint (e.g., one task is to classify different types of animals and another task is to classify different breeds of dogs). Formally, let $\mathcal{Y}_t \subseteq \mathcal{Y}$ (the label space of all tasks learned so far).  The goal of TIL is to learn a function $f: \mathcal{X} \times \mathcal{T} \to \mathcal{Y}$ .

In a typical TIL approach, a \textit{multi-head} configuration is applied, where each task is allocated an exclusive output layer or head. 
Since the task-ID is given for each test sample, the task-specific model is applied in classification. Since many TIL approaches has achieved no forgetting, the main challenge of TIL now is bi-directional (forward and backward)  knowledge transfer.




\textbf{(2) Domain-incremental learning (DIL).} In this setting, the class labels are assumed to be the same across all tasks (the inputs are from different domains). A sequence of sentiment classification tasks can be regarded as a DIL problem because all tasks have the same three class labels, \textit{positive}, \textit{negative}, and \textit{neutral}. Generation tasks in NLP also form a DIL problem as the LM head has the same number of ``classes'' (vocabulary tokens). Formally, let $\mathcal{Y}$ be the set of class labels for every task. 
{The goal of DIL is to learn a function: $f: \mathcal{X} \times \mathcal{T} \to \mathcal{Y}$.} 


Some existing
DIL systems do not use task-ID in testing, {i.e.}, $f: \mathcal{X} \to \mathcal{Y}$, and can still work well. This is because (1) the tasks are very similar (e.g., sentiment classification tasks that do not have many conflicts) \cite{ke2021Classic}, (2) the tasks are very dissimilar and  the domain/task of a test sample can be easily  predicted~\cite{madotto2020continual,gururangan2021demix}, and (3) there is a strong replay module~\cite{DBLP:conf/iclr/QinJ22}.  {\color{black}To distinguish the two cases, we use DIL (w/ ID) and DIL (w/o ID) to refer to DIL with and without task-ID, respectively.}

A typical DIL method uses a \textit{single-head} configuration, which causes additional task interference in training. While most papers regard DIL as a separate setting, it is really a special case of TIL (see their definitions above). A DIL problem can always be solved with a TIL method. Like TIL, in general task-ID is needed for each test sample because different domains can be different and may even have conflicts. For example, the same sentence ``\textit{This unit is sucks}'' is positive for a vacuum cleaner but negative for most other products or domains. 

\textbf{(3) Class-incremental learning (CIL)}. CIL assumes no task-ID is provided at test time. CIL is mainly used for supervised end-task learning, where each task has a set of non-overlapping classes. Only a single model is built for all classes learned so far. In testing, a test instance from any class may be presented for classification. Formally, we have 
 $\mathcal{Y}_t \cap \mathcal{Y}_{t'}= \varnothing, \forall~t \ne t'$ and $\mathcal{Y}=\bigcup_{t=1}^T \mathcal{Y}_t$. 
 The goal of CIL is to learn a function $f: \mathcal{X} \to \mathcal{Y}$. 

A typical CIL approach uses a \textit{single-head} configuration to cover all classes. CIL is more difficult than TIL and DIL because it has an additional challenge (apart from CF and KT). It is called \textbf{\textit{inter-task class separation}} (ICS), which is identified recently in \cite{kim2022theoretical}. 
ICS means that when learning a new task, since the system does not see the data of the previous tasks, it cannot establish decision boundaries between the classes in the current task and the classes in previous tasks. Given that at test time no task-ID is given, CIL accuracy is often much lower than that of TIL or DIL. We will discuss how this problem may be solved in Sec.~\ref{sec.appro_ts}. 


\textbf{Focus of NLP.} In ML and CV research, almost only TIL and CIL are studied. However, in NLP, all settings have been used due to NLP's rich types of tasks. Further, CL research in ML and CV mainly focuses on overcoming CF. In NLP, bi-directional KT is also extremely important. That is because (1) in text, words and phrases in different tasks or domains normally have the same meaning and (2) many NLP tasks are similar and have shared knowledge, e.g., different information extraction tasks and different sentiment classification tasks. 








\begin{table*}[h]
\centering
\resizebox{0.8\textwidth}{!}{
\begin{tabular}{ccc||ccc}
\specialrule{.2em}{.1em}{.1em}
CL Settings & Type & Mode & NLP Problems & CL Papers  \\
\specialrule{.1em}{.05em}{.05em}
\multirow{7}{*}{TIL} & \multirow{7}{*}{End-task} & \multirow{7}{*}{Offline} & \multirow{2}{*}{Aspect sentiment classification}  & B-CL~\cite{ke2021adapting} \\
& & & & CTR~\cite{ke2021achieving} \\
 \cline{4-5}
  & & &  \multirow{2}{*}{Intent classification} & MeLL~\cite{DBLP:conf/kdd/0001PLCQZHCL021}  \\
   & & & & PCLL~\cite{DBLP:journals/corr/abs-2210-07783}  \\
\cline{4-5}
 &  &&  Slot filling & PCLL~\cite{DBLP:journals/corr/abs-2210-07783}\\
 &  & &Topic classification & CTR~\cite{ke2021achieving}\\
  &  & &Mixed diverse tasks & CLIF~\cite{DBLP:conf/emnlp/JinLR021} \\
\cline{1-5}
\multirow{12}{*}{DIL (w/o ID)} & \multirow{12}{*}{End-task} & \multirow{12}{*}{Offline} & Mixed 5 classification tasks & LFPT5~\cite{DBLP:conf/iclr/QinJ22} \\
& & & Named-entity recognition &  LFPT5~\cite{DBLP:conf/iclr/QinJ22} \\
& & & Summerization & LFPT5~\cite{DBLP:conf/iclr/QinJ22}  \\
& & & Paraphrase & RMR-DSE~\cite{li-etal-2022-overcoming} \\
\cline{4-5}
& & & \multirow{2}{*}{Dialogue response generation} & RMR-DSE~\cite{li-etal-2022-overcoming} \\
& & & & AdapterCL~\cite{madotto2020continual} \\
\cline{4-5}
& & & Dialogue state tracking & AdapterCL~\cite{madotto2020continual}  \\
& & & Dialogue end2end  & AdapterCL~\cite{madotto2020continual} \\
& & & Aspect sentiment classification & CLASSIC~\cite{ke2021Classic}  \\
\cline{3-5}
& & \multirow{2}{*}{{\begin{tabular}[c]{@{}c@{}}Online \\ (Blurry)\end{tabular}}} & \multirow{2}{*}{Question answering} & MBPA++~\cite{DBLP:conf/nips/dAutumeRKY19} \\
& & & & Meta-MBPA++~\cite{wang2020efficient}  \\
\cline{2-5}
& \multirow{3}{*}{DP} & \multirow{3}{*}{Offline} & 5 pre-training domains & ELLE~\cite{qin2022elle}  \\
 & & & {8 pre-training domains} & {DEMIX~\cite{gururangan2021demix}}\\
 & & & 10 pre-training domains & Continual-T0~\cite{scialom2022continual}\\
 \cline{1-5}
\multirow{7}{*}{DIL (w/ ID)} & \multirow{6}{*}{End-task} & \multirow{6}{*}{Offline} & Mixed 5 classification tasks & LAMOL~\cite{sun2020lamol}\\
&  & &  Mixed classification and labeling tasks & LAMOL~\cite{sun2020lamol}  \\
& & & Dialogue state tracking & C-PT~\cite{zhu2022continual} \\
&  & & Dialogue response generation  & TPEM~\cite{DBLP:conf/acl/GengYXSX020}  \\
& & & Mixed classification and generation & ConTinTin~\cite{DBLP:conf/acl/0001LX22}  \\
& & & Mixed 4 generation tasks & ACM~\cite{zhang2022continual} \\
\cline{2-5}
& DP &Offline & 5 pre-training domains & CPT~\cite{ke2022cpt}  \\
\cline{1-5}
\multirow{12}{*}{CIL}  & \multirow{8}{*}{End-task} & \multirow{6}{*}{Offline} & Named-entity recognition & ExtendNER~\cite{DBLP:conf/aaai/MonaikulCFR21} \\
\cline{4-5}
& & & \multirow{2}{*}{Intent classification} & CID~\cite{DBLP:journals/corr/abs-2108-04445} \\
& & & & PAGeR~\cite{DBLP:conf/naacl/VarshneyPKVS22} \\
\cline{4-5}
& & & Mixed 5 classification tasks & IDBR~\cite{huang2021continual}  \\
& & & Slot filling & ProgM~\cite{shen-etal-2019-progressive}\\
& & &  Sentence representation & SRC~\cite{DBLP:conf/naacl/LiuUS19} \\
\cline{3-5}
&  & \multirow{2}{*}{\begin{tabular}[c]{@{}c@{}}Online \\ (Hard)\end{tabular}} & \multirow{2}{*}{Mixed 5 classification tasks} & MBPA++~\cite{DBLP:conf/nips/dAutumeRKY19}  \\
& & & & Meta-MBPA++~\cite{wang2020efficient}  \\
\cline{2-5}
& \multirow{2}{*}{Few-shot} & \multirow{2}{*}{Offline} & \multirow{2}{*}{Intent classification} & ENTAILMENT~\cite{DBLP:conf/naacl/XiaYFY21}\\
& & & & CFID~\cite{DBLP:conf/coling/LiZCGZZ22} \\
\specialrule{.1em}{.05em}{.05em}
\end{tabular}
}
\vspace{-3mm}
\caption{CL tasks and NLP problems. The same datasets may be formulated for different CL settings, depending on how a task is defined. ``DP'' stands for domain-adaptive pre-training. ``w/ ID'' and ``w/o ID'' stand for with and without ID correspondingly. 
}
\label{tab:nlp_to_cl}
\vspace{-4mm}
\end{table*}

\subsection{Learning Modes}

Learning modes control how the system is trained. They may be used with any of the aforementioned settings. In the literature, there are two modes:

\textbf{(1) Offline or Batch CL.} When a task arrives, all the training data of the task is available and the training can be done in any number of epochs.

\textbf{(2) Online CL}~\cite{mai2022online}. In this mode, the training data are assumed to come in a data stream. Whenever a small batch of data is accumulated in the stream, it is trained in one iteration. So online CL trains only in one epoch. Existing online CL research mainly solves the CIL problem. Online CL has two task arriving configurations. 

\textbf{(a) Hard task boundary}~\cite{DBLP:conf/nips/dAutumeRKY19}. In this configuration, the data of each task comes in a stream continuously without any data in between from any other tasks. That is, the data of another task will only come after the data from its preceding task has been exhausted. 

\textbf{(b) Blurry task boundary}~\cite{DBLP:conf/nips/AljundiLGB19}. In this mode, the data from different tasks may intertwine. Thus, the boundaries of tasks are blurry.  

\section{NLP Tasks in CL} 

This section summarizes NLP problems that have been solved as CL tasks. Due to different natures of the fields, NLP and CV have different CL tasks. 
Below, we first introduce different task types and then present a task summary. 

\subsection{Types of Tasks}
\label{sec.task_type}

A task in CL can be of different types. We classify them into three types:

\textbf{(1) End-task}. This is the most popular task type where each task is an end-task, which can be any NLP problem such sentiment classification, named entity recognition or summarization. 

\textbf{(2) Few-shot end-task}~\cite{DBLP:conf/cvpr/TaoHCDWG20}. Each task is a few-shot learning task. Since it is hard to learn a few-shot task well due to possible over-fitting, the systems need to address both the few-shot learning and continual learning issues.

\textbf{(3)~Domain-adaptive pre-training task}~\cite{ke2022cpt}. Each task uses a domain corpus to adapt or fine-tune a language model (LM) to the specific domain. The LM may be trained from scratch or from an existing pre-trained checkpoint. 



\subsection{CL Formulation of NLP Problems}

We summarize CL formulation of NLP problems in the existing literature in Table~\ref{tab:nlp_to_cl}. Due to the fact that a large number of CL papers have been published, the list is by no means exhaustive.

We observe that a large number of NLP problems belong to DIL. This is very different from CV where only a few problems are formulated as DIL. This is mainly because a language model (LM) is naturally a text generator and recent works on large LMs rely on the intuition that most NLP tasks can be described via natural language instructions (see Appendix~\ref{sec.preliminary}). With the instructions, different types of tasks (e.g., classification or extraction) can be converted to generation tasks and thus a shared LM head is applied across all tasks.

\section{Approaches for Continual Learning}
\label{sec.approch}

In this section, we give a comprehensive review of the existing CL techniques. It is important to keep in mind that different settings have different challenges and thus different approaches. However, since all settings have to deal with the challenges of \textit{catastrophic forgetting} (CF) and \textit{knowledge transfer} (KT), it is no surprise that some techniques have been used in different problem settings. 
We categorize existing systems based on their family of approaches and also give the setting that each system belongs to. We first discuss the techniques for CF prevention, and then extend the techniques to also include bi-directional (forward and backward) KT. Lastly, we introduce the latest challenge of \textit{inter-task class separation} (ICS). 

We note that the CF prevention is closely related to \textit{stability-plasticity} dilemma where the main concern is to balance the network stability (preserving the learned knowledge or preventing CF) and plasticity (accurately learning the new experience or task). In contrast, bi-directional KT is an extended view which considers both CF and KT in both forward and backward directions. 
We also note that one system may combine more than one family of approaches. To further benefit the NLP community, we also discuss how the NLP-based CL systems are similar or different from CV-based CL systems.
Due to space limits, \textbf{details of each system} are given in Appendices~\ref{sec.sys_details_cf}, \ref{sec.sys_details_kt}, and \ref{sec.sys_details_tsf}.

\subsection{Approaches for CF Prevention}
\label{sec.appro_cf}

The main idea of CF prevention is to reduce the degree of parameter-sharing among tasks or to prevent major changes to previously learned parameters. Since the system also needs to consider plasticity, it has to use different strategies to balance what to retain and what to use for new tasks.

We categorize existing techniques for overcoming CF into  \textit{regularization-based} (Sec.~\ref{sec.reg_cf}), \textit{replay-based} (Sec.~\ref{sec.replay}) and \textit{parameter-isolation based} (Sec.~\ref{sec.parameter_isolate}) families. In addition, we add the \textit{instruction based} (Sec.~\ref{sec.instruct_cf}) family specifically for NLP. A summary can be found in Table~\ref{tab:cf_prevention}.  

\begin{table*}[t]
\centering
\resizebox{0.85\textwidth}{!}{
\begin{tabular}{c|ccccc}
\specialrule{.2em}{.1em}{.1em}
Desiderata & Family (-based) & Sub-family (-based) & Settings & CL Papers  \\
\specialrule{.1em}{.05em}{.05em}
\multirow{35}{*}{CF Prevention} & \multirow{10}{*}{Regularization} & \multirow{3}{*}{Importance} & \multirow{2}{*}{DIL (w/o ID)}  & RMR-DSE \cite{li-etal-2022-overcoming} \\
& & & & AEWC~\cite{DBLP:journals/corr/abs-1712-09943} \\
 \cline{4-5}
  & & & CIL & SRC~\cite{DBLP:conf/naacl/LiuUS19}  \\
\cline{3-5}
 &  & \multirow{5}{*}{Distillation} &  DIL (w/o ID) & LFPT5~\cite{DBLP:conf/iclr/QinJ22}\\
  \cline{4-5}
 &  &  & \multirow{4}{*}{CIL} & ExtendNER~\cite{DBLP:conf/aaai/MonaikulCFR21}\\
  &  & & & CFID~\cite{DBLP:conf/coling/LiZCGZZ22} \\
 &  & &  & CID~\cite{DBLP:journals/corr/abs-2108-04445}\\
&  & &  & PAGeR~\cite{DBLP:conf/naacl/VarshneyPKVS22}  \\
\cline{2-5}
& \multirow{10}{*}{Replay} & \multirow{5}{*}{{\begin{tabular}[c]{@{}c@{}}Raw sample\\(Training)\end{tabular}}} & \multirow{3}{*}{CIL} & CFID~\cite{DBLP:conf/coling/LiZCGZZ22} \\
 &  & &  & CID~\cite{DBLP:journals/corr/abs-2108-04445}\\
  &  & &  & IDBR~\cite{huang2021continual}\\
   \cline{4-5}
&  & & \multirow{2}{*}{DIL (w/o ID; DP)} & Continual-T0~\cite{scialom2022continual}  \\
&  & &  & ELLE~\cite{qin2022elle} \\
   \cline{3-5}
  & & \multirow{2}{*}{{\begin{tabular}[c]{@{}c@{}}Raw sample\\(Training and testing)\end{tabular}}} & CIL (online) & MBPA++~\cite{DBLP:conf/nips/dAutumeRKY19} \\
& & & DIL (w/o ID; online) & Meta-MBPA++~\cite{wang2020efficient} \\
\cline{3-5}
& & Prototype & TIL & MeLL~\cite{DBLP:conf/kdd/0001PLCQZHCL021} \\
\cline{3-5}
& & \multirow{5}{*}{Pseudo-sample} & TIL & PCLL~\cite{DBLP:journals/corr/abs-2210-07783} \\
\cline{4-5}
& &  &\multirow{2}{*}{DIL (w/ ID)}  & LAMOL~\cite{sun2020lamol} \\
& &  &  & ACM~\cite{zhang2022continual} \\
\cline{4-5}
& & & DIL (w/o ID) & LFPT5~\cite{DBLP:conf/iclr/QinJ22} \\
& & & CIL & PAGeR~\cite{DBLP:conf/naacl/VarshneyPKVS22} \\
\cline{2-5}
& \multirow{12}{*}{Parameter-isolation} & \multirow{4}{*}{Task masking} & \multirow{2}{*}{TIL} & CTR~\cite{ke2021achieving}  \\
& &  &  & B-CL~\cite{ke2021adapting} \\
\cline{4-5}
& &  & DIL (w/ ID; DP) & CPT~\cite{ke2022cpt} \\
& &  & DIL (w/o ID) & CLASSIC~\cite{ke2020continual} \\
\cline{3-5}
& & Parameter generation & TIL & CLIF~\cite{DBLP:conf/emnlp/JinLR021} \\
\cline{3-5}
& & \multirow{6}{*}{Dynamic architecture} & \multirow{2}{*}{DIL (w/ ID)} & C-PT~\cite{zhu2022continual} \\
& & & & TPEM~\cite{DBLP:conf/acl/GengYXSX020} \\
\cline{4-5}
& & & \multirow{2}{*}{DIL (w/o ID; DP)} &  DEMIX~\cite{gururangan2021demix} \\
& & & & ELLE~\cite{qin2022elle} \\
\cline{4-5}
& & & DIL (w/o ID) & AdapterCL~\cite{madotto2020continual} \\
& & & CIL & ProgM~\cite{shen-etal-2019-progressive} \\
\cline{2-5}
& \multirow{2}{*}{Instruction} & \multirow{2}{*}{N/A} & DIL (w/ ID) & ConTinTin~\cite{DBLP:conf/acl/0001LX22}  \\
\cline{4-5}
& & & CIL & ENTAILMENT~\cite{DBLP:conf/naacl/XiaYFY21} \\

\specialrule{.1em}{.05em}{.05em}
\end{tabular}
}
\vspace{-2mm}
\caption{
Families of approaches for CF prevention in different NLP systems. ``DP'' stands for domain-adaptive pre-training. ``w/ ID'' and ``w/o ID'' stand for with and without ID correspondingly. 
}
\label{tab:cf_prevention}
\vspace{-4mm}
\end{table*}



\subsubsection{Regularization-based Methods}
\label{sec.reg_cf}

The main idea of this family is to add a penalty or regularization to the loss function to penalize changes to important parameters learned from previous tasks in learning a new task. The main drawback is that it introduces a trade-off between the new task learning and forgetting prevention.
This approach was originally proposed in CV~\cite{Kirkpatrick2017overcoming,zenke2017continual,He2018overcoming,DBLP:conf/iclr/EbrahimiEDR20} and has been adopted by the NLP community. 

\textbf{(1). Regularizing the loss based on parameter importance}~\cite{li-etal-2022-overcoming,DBLP:journals/corr/abs-1712-09943,DBLP:conf/naacl/LiuUS19}. The most popular and established regularization-based method in this sub-family is EWC~\cite{Kirkpatrick2017overcoming} in CV. Note that updating a pre-trained LM in CL will inevitably result in forgetting of some knowledge in the LM in this sub-family. This is because the data used in the pre-training process is not accessible to end-users of the LM. Then parameter importance is not compute-able by end-users in order to protect the important LM parameters. As a result, many systems fix the LM and train only some additional modules {\color{black}(e.g., adapter or prompt) added to the LM.}

\textbf{(2). Regularizing the loss based on distillation}~\cite{DBLP:conf/aaai/MonaikulCFR21,DBLP:conf/coling/LiZCGZZ22,DBLP:journals/corr/abs-2108-04445,DBLP:conf/naacl/VarshneyPKVS22,DBLP:conf/iclr/QinJ22}. This sub-family distills the knowledge from a previous model (trained on a previous task) to the model being trained on the new data so that the previous knowledge is not forgotten. 
The main drawback is that  distillation is vulnerable to  domain shift between tasks \cite{Aljundi2016expert}.

\textbf{Other regularization-based methods.} Though existing NLP systems only regularize the loss, CV-based systems also regularize the gradient~\cite{zeng2019continuous,Wang_2021_CVPR,DBLP:conf/iclr/EbrahimiEDR20,DBLP:conf/nips/MirzadehFPG20}.

\subsubsection{Replay-based Methods}
\label{sec.replay}

Most replay-based methods either (a) store a small subset of training samples of previous tasks
in a memory or (b) learn a data generator to generate pseudo samples of previous tasks. (b) is usually called \textit{pseudo/generative replay}. In learning a new task, the saved samples or the generated samples and the new task data are both used in training~\cite{Rebuffi2017,Lopez2017gradient,Shin2017continual,Kemker2018fearnet}. 

\textbf{(1). Replaying raw samples in training}~\cite{DBLP:conf/coling/LiZCGZZ22,DBLP:journals/corr/abs-2108-04445,scialom2022continual,qin2022elle,huang2021continual}. This sub-family directly applies the case (a) above.  

\textbf{(2). Replaying raw samples in training and inference}~\cite{DBLP:conf/nips/dAutumeRKY19,wang2020efficient}. Besides joint training of both the replay samples and new samples, this approach also uses the replay samples in inference for local adaptation (re-training the model using the closest replay data to the testing sameple).

\textbf{(3). Replaying prototypes.} Instead of storing raw samples, one can also save class prototypes~\cite{DBLP:conf/kdd/0001PLCQZHCL021}.

\textbf{(4). Optimizing on the replay memory.} Using the replay samples has the risk of over-fitting. This sub-family 
uses the saved data only to regularize the training, but does not train the replay samples. Representative systems are GEM~\cite{Lopez2017gradient} and A-GEM~\cite{Chaudhry2019ICLR}. Both systems use an inequality constraint to prevent parameter update from increasing the loss of each previous task. The previous task loss is estimated by using samples in the memory. 

\textbf{(5). Replaying generated pseudo samples.} This is the case (b). Some NLP systems leverage  LMs to generate pseudo samples for previous tasks~\cite{DBLP:journals/corr/abs-2210-07783,DBLP:conf/naacl/VarshneyPKVS22,sun2020lamol,zhang2022continual,DBLP:conf/iclr/QinJ22}. This approach is becoming popular because an LM is naturally a generator and can generate high-quality pseudo samples. While this is promising, it has been observed that the LM has difficulty to generate samples for some tasks, such as summarization~\cite{DBLP:conf/iclr/QinJ22} and aspect sentiment classification~\cite{ke2021achieving}. 

\textbf{(6). Replaying features (latent representations).} Generating high-quality samples can be very challenging. Some CV papers proposed to generate features. Example systems are those in \cite{van2020brain,DBLP:conf/eccv/YeB20}.


\begin{table*}[h]
\centering
\resizebox{0.8\textwidth}{!}{
\begin{tabular}{c|ccccc}
\specialrule{.2em}{.1em}{.1em}
Desiderata & Family (-based) & Sub-family (-based) & Settings & CL Papers  \\
\specialrule{.1em}{.05em}{.05em}
\multirow{10}{*}{Bi-directional KT} & \multirow{4}{*}{Similarity} & Importance & DIL (w/ ID)  & ACM~\cite{zhang2022continual} \\
 \cline{3-5}
& & \multirow{2}{*}{Feature (CapsNet)}  & \multirow{2}{*}{TIL} & CTR~\cite{ke2021achieving}  \\
& &  & & B-CL~\cite{ke2021adapting}  \\
\cline{3-5}
& & Feature (Contrast)  & DIL (w/o ID) & CLASSIC~\cite{ke2020continual}  \\
  \cline{2-5}
& \multirow{6}{*}{Replay} & Regularization  & CIL & IDBR~\cite{huang2021continual}  \\
  \cline{3-5}
& & \multirow{3}{*}{Meta-Learning}  & TIL & MeLL~\cite{DBLP:conf/kdd/0001PLCQZHCL021}  \\
& &  & DIL (w/o ID; online) & Meta-MBPA++~\cite{wang2020efficient} \\
& &  & CIL (online) & MBPA++~\cite{DBLP:conf/nips/dAutumeRKY19} \\
 \cline{3-5}
& & \multirow{2}{*}{Instruction} & \multirow{2}{*}{DIL (w/ ID)} & C-PT~\cite{zhu2022continual}  \\
& & & & ConTinTin~\cite{DBLP:conf/acl/0001LX22}  \\
\specialrule{.1em}{.05em}{.05em}
\end{tabular}
}
\vspace{-2mm}
\caption{
Families of approaches for bi-directional KT in different NLP systems. ``w/ ID'' and ``w/o ID'' stand for with and without ID correspondingly. 
}
\label{tab.bidirection_kt}
\vspace{-4mm}
\end{table*}

\subsubsection{Parameter-isolation Based Methods}
\label{sec.parameter_isolate}
\textit{Parameter-isolation} based methods allocate different parameters to different tasks to prevent subsequent tasks from interfering with the previously learned parameters.
{\color{black}This family requires task-ID in both training and testing so it is mainly for TIL and DIL settings.} 
This family also usually suffers from the learning capacity problem due to masked neurons or parameters and makes KT challenging. Like regularization-based and replay-based methods, parameter isolation methods are also originated from CV~\cite{DBLP:journals/corr/RusuRDSKKPH16,yoon2018lifelong,fernando2017pathnet,Mallya2017packnet}. 

\textbf{(1). Fixed architecture.} This sub-family isolates a subset of parameters for each task in a fixed network. 
Three popular methods are \textit{task masking}, \textit{parameter generation} and \textit{sub-network masking}.

\textbf{(i). Task masking.} This approach 
masks a subset of neurons at each layer for a task (identified by task-ID). Since the mask can indicate what neurons have been used by previous tasks, in learning the new task, the system can freeze the used neurons to prevent CF. 
The most popular system is HAT~\cite{Serra2018overcoming} in CV. The details of HAT is in Appendix~\ref{sec.paramter_isolate}.
In NLP, this task masking mechanism has been mainly used in the adapter layer to prevent CF~\cite{ke2021achieving,ke2021adapting,ke2021Classic,ke2022cpt}.

\textbf{(ii). Parameter generation.} This method uses one network to generate the model parameters of another network. Since the generation can be conditioned on the task-ID, it helps mitigate CF.  Hypernet~\cite{von2019continual} in CV and CLIF~\cite{DBLP:conf/emnlp/JinLR021} in NLP take this approach. 

\textbf{(iii). Subnetwork masking.} {\color{black}This method fixes a network $N$} and trains another mask network $M$ to find a sub-network in $N$ for each task. An example in CV is SupSup~\cite{wortsman2020supermasks}.

\textbf{(2). Dynamic architecture}~\cite{gururangan2021demix,qin2022elle,zhu2022continual,DBLP:conf/acl/GengYXSX020,madotto2020continual,shen-etal-2019-progressive}. This sub-family expands the network for each new task.

\textbf{(3). Parameter pool.} This approach initializes a pool of parameters and does parameter selection. L2P~\cite{wang2021learning} in CV is an example.

\subsubsection{Instruction-based Methods}
\label{sec.instruct_cf}

This family is specific to NLP. 
It uses task-specific instructions to condition the LM.  
An instruction is a few words. The conditioning may not be sufficient to prevent CF. So far, it has been used only DIL and CIL systems~\cite{DBLP:conf/acl/0001LX22,DBLP:conf/naacl/XiaYFY21}.

\subsection{Approaches for Knowledge Transfer} %
\label{sec.appro_kt}

We now review approaches for knowledge transfer (KT). KT requires the system to not only consider CF, but also bi-directional transfer. Clearly, simply reducing parameter-sharing as in CF prevention cannot help encourage transfer. KT also needs 
to focus on  deciding what used parameters should be allowed to update (backward and forward KT) and/or to share (forward KT). The core idea for KT can be separated into two families, one that leverages the similarity among tasks (Sec.~\ref{sec.similarity}) and the other that leverages the replay data (Sec.~\ref{sec.replay}). A summary can be found in Table~\ref{tab.bidirection_kt}, and more details about each system are in Appendix~\ref{sec.sys_details_kt}.

\subsubsection{Similarity-based Methods}
\label{sec.similarity}

The idea is that two tasks can transfer knowledge to each other \textit{if and only} if they are similar. KT is typically achieved by sharing some parameters by similar tasks.
Since the similarity is not provided by the task data, this family often uses a proxy to compute task similarity. Depending on the proxy, 3 sub-families have been proposed in the literature.

\textbf{(1). Importance-proxy.} 
This approach first looks for the previous knowledge that is important for the current task and then trains the detected important knowledge together with the current task. The most naive way is to train a mixing coefficient to detect the importance~\cite{zhang2022continual}.


\textbf{(2). Feature-proxy.} 
This approach compares the feature from previous task models with the feature for the current task. There are two main ideas. 

\textbf{(i). Capsule Network (CapsNet).} 
A simple \textit{capsule network (CapsNet)}~\cite{hinton2011transforming,sabour2017dynamic} consists of two capsule layers. The first layer stores low-level feature maps, and the second layer generates the classification probability with each capsule corresponding to one class. CapsNet uses a \textit{dynamic routing} algorithm to make each lower-level capsule send its output to a similar or ``agreed'' {\color{black}(computed by dot product)} higher-level capsule to enable KT. This has been adopted in several NLP systems~\cite{ke2021achieving,ke2021adapting}


\textbf{(ii). Contrastive learning}~\cite{ke2021achieving}. The idea is to leverage contrastive learning to capture the shared knowledge between different views. If one can generate a view from previous tasks that is similar to the current task, the contrastive loss can capture the shared knowledge and learn a representation for KT to the new task learning.

\textbf{(3). Gradient-proxy.} Gradient space is another possible proxy for task similarity computation.
The idea is that if the new task's gradients are \textit{sufficiently} aligned with an old task's subspace, the updating of the learned model is likely to improve its performance.
However, what is ``sufficient'' is decided by thresholds which can be inaccurate. This is only used in CV so far~\cite{DBLP:journals/corr/abs-2211-00789,lin2022trgp}


\begin{table*}[h]
\centering
\resizebox{0.8\textwidth}{!}{
\begin{tabular}{c|cccc}
\specialrule{.2em}{.1em}{.1em}
Challenge & Family (-based)  & Settings & CL Papers  \\
\specialrule{.1em}{.05em}{.05em}
\multirow{12}{*}{Inter-task Class Separation} & \multirow{8}{*}{Replay}  & \multirow{5}{*}{CIL} & IDBR~\cite{huang2021continual}  \\
& & & ExtendNER~\cite{DBLP:conf/aaai/MonaikulCFR21}\\
& & & CFID~\cite{DBLP:conf/coling/LiZCGZZ22} \\
&  & & CID~\cite{DBLP:journals/corr/abs-2108-04445}\\
 &  & & PAGeR~\cite{DBLP:conf/naacl/VarshneyPKVS22}  \\
  \cline{3-4}
& & DIL (w/o ID) & LFPT5~\cite{DBLP:conf/iclr/QinJ22} \\
\cline{3-4}
& & DIL (w/o ID; online) & Meta-MBPA++~\cite{wang2020efficient} \\
&  & CIL (online) & MBPA++~\cite{DBLP:conf/nips/dAutumeRKY19} \\
\cline{3-4}
&  & \multirow{2}{*}{DIL (w/o ID; DP)} & Continual-T0~\cite{scialom2022continual}  \\
 & & & ELLE~\cite{qin2022elle} \\
\cline{2-4}
& \multirow{3}{*}{Post-prediction} & \multirow{2}{*}{DIL (w/o ID)} & AdapterCL~\cite{madotto2020continual} \\
 & &  & RMR-DSE \cite{li-etal-2022-overcoming} \\
   \cline{3-4}
 & & DIL (w/o ID; DP) & DEMIX~\cite{gururangan2021demix} \\
\specialrule{.1em}{.05em}{.05em}
\end{tabular}
}
\vspace{-2mm}
\caption{
Families of approaches to solving ICS in CIL (or task separation in DIL) in different NLP systems. ``DP'' stands for domain-adaptive pre-training. ``w/ ID'' and ``w/o ID'' stand for with and without task-ID respectively. 
}
\label{tab.task_sep}
\vspace{-4mm}
\end{table*}





\subsubsection{Replay-based Methods}

{\color{black}In CF prevention, the replay data is used to minimize the backward interference. In contrast, the replay data for KT is used to learn transferable knowledge. Depending on how to achieve it, three sub-families are proposed.} 

\textbf{(1) Regularization-based}~\cite{huang2021continual}.
Instead of penalizing the change of parameters, regularization for KT disentangles task-shared knowledge and task-specific knowledge.

\textbf{(2) Meta-learning-based}~\cite{DBLP:conf/kdd/0001PLCQZHCL021,wang2020efficient,DBLP:conf/nips/dAutumeRKY19}.
Meta-learning is known to be able to capture the transferable knowledge as it learns a large set of tasks. The main idea of this family is to leverage this property to maximize transfer and minimize interference.
To enable KT, the key is to make the support set contain only the new task while the query set contain examples drawn from all learned tasks. With this meta-objective design, different meta-representations can be trained for continual learning. Details are given in Appendix~\ref{sec.meta}.

\textbf{(3) Instruction-based}~\cite{zhu2022continual,DBLP:conf/acl/0001LX22}.
Apart from being used to address CF (Sec.~\ref{sec.instruct_cf}), instructions have also been used to enable KT. The idea is to give an instruction that is similar or related to previous tasks. 

\subsection{Dealing with Inter-task Class Separation}
\label{sec.appro_ts}


In Sec.~\ref{sec.setttings}, we discussed the additional challenge of \textit{inter-task class separation} (ICS) for CIL (beyond CF and KT) \cite{kim2022theoretical}, which refers to the difficulty to establish decision boundaries between the classes of the current task and those of the previous tasks in CIL because in learning the current task, the learner does not see the previous task data. \cite{kim2022theoretical} theoretically shows that good \textit{within-task prediction} (WP) (predicting only the classes within each task like that in TIL) and OOD detection for each task are \textit{necessary and sufficient conditions} for strong CIL performances. Based on the theory, this paper proposed a technique that combines a TIL method that can overcome CF and a OOD detection method to build the model for each task. Note that most OOD method can also classify in-distribution data, which is WP. This approach achieves the SOTA performance in CIL. The following approaches are also used to deal with ICS. 
A summary is in Table~\ref{tab.task_sep}.

\textbf{Replay-based methods}~\cite{huang2021continual,DBLP:conf/aaai/MonaikulCFR21,DBLP:conf/coling/LiZCGZZ22,DBLP:journals/corr/abs-2108-04445,DBLP:conf/naacl/VarshneyPKVS22,DBLP:conf/iclr/QinJ22,wang2020efficient,DBLP:conf/nips/dAutumeRKY19,scialom2022continual,qin2022elle} partially deal with the ICS problem because the learner can see some past data, but since the amount of replay data is  small, the ICS problem is not solved. 

\textbf{Post-prediction methods}~\cite{madotto2020continual,gururangan2021demix,li-etal-2022-overcoming}. This is mainly used in DIL systems. Sec.~\ref{sec.setttings} mentioned that some DIL systems do not provide task-ID in testing, which thus faces the same ICS problem. However, since the classes of all tasks are the same in DIL, we do not separate the classes, but separate the domains. In this case, we can regard each domain as a virtual class and the same theory still applies. 
Without  dealing with the ICS problem, some heuristic methods are proposed to \textit{post-predict} the domain for each test sample \textit{after training}.
They may perform well mainly due to the specific problem property (e.g., the domains are very dissimilar and easy to differentiate). {\color{black}\textit{Continual domain-adaptive pre-training} of LMs involving fine-tuning for end-tasks after pre-training without providing domain-ID also belongs to this sub-family.}


\section{Observations and Future Directions}

Due to space limits, CL evaluation is put in Appendix~\ref{sec.evaluate}. This section discusses some observations about existing methods. Based on these observations, we outline some future research directions.

\textbf{Knowledge transfer is the major issue in TIL.} Several TIL methods have achieved no CF. Maximizing KT is still challenging \cite{ke2021achieving,ke2021adapting,DBLP:conf/kdd/0001PLCQZHCL021,zhu2022continual}. Although Sec.~\ref{sec.appro_kt} discussed some systems that perform KT, the amount of transfer is still limited. 
The main issue is that the task sequence can contain a mix of similar and dissimilar tasks (e.g., CAT~\cite{ke2020mixed}) and it is hard to detect what tasks are similar and have shared knowledge and what tasks are dissimilar and cannot share. If such a detection is wrong, serious CF can arise, especially for backward transfer. More research is still needed.   

\textbf{CF and ICS are still major issues for CIL and DIL.} Although \cite{kim2022theoretical} offered a approach to deal with ICS through OOD detection, the accuracy results are still far below those from training all tasks/classes together (upper bound). Much more future research is still needed. 

\textbf{Continual domain-adaptive pre-training are still in their infancy.} A domain adaptively pre-trained LM can improve end-tasks, but how to maintain the needed general knowledge (GK) already in the LM is challenging because learning one task by fine-tuning the LM can cause serious CF for the GK originally in the LM and make the LM unsuitable for subsequent tasks.  
Although some initial attempts have been made~\cite{gururangan2021demix,ke2022cpt,scialom2022continual,qin2022elle}, they are very limited. 

\textbf{Scalablity is an issue.} When a very large number of tasks are learned, the capacity of a network will be a problem. Most  CL systems learn a small number of tasks ($<100$) and may also expand the network (e.g., \cite{DBLP:journals/corr/RusuRDSKKPH16,madotto2020continual}). Some preliminary works dealing with the capacity issue include distillation~\cite{DBLP:conf/icml/Schwarz0LGTPH18},  pruning~\cite{DBLP:conf/nips/Hung0WCCC19} and regularization~\cite{Serra2018overcoming}, but it is still unclear how the capacity issue affects the performance and how to alleviate the issue effectively.

\textbf{Limited CL work on Temporal CL.} In this case, we have only one task, but the task needs to be constantly updated as the time goes. 
So far, limited research has been done except  
some analysis and dataset papers 
\cite{luu2021time,rempel2014mind,dhingra2022time,loureiro2022timelms,rottger2021temporal}.

\section{Conclusion}
\vspace{-1mm}
{\color{black}An AI agent is not truly intelligent if it does not have the continual learning (CL) capability to learn and to accumulate knowledge over time and to use the learned knowledge to enable better future learning.} This paper gave a comprehensive survey of the recent advances of CL in NLP and discussed some 
future directions. We hope that this survey will inspire NLP researchers to design better algorithms. 


\section{Ethics Statement}

This work presents a comprehensive literature survey of the state-of-the-art continual learning research in NLP. As more and more chatbots and other AI agents with natural language understanding capabilities appear in our lives, we believe that leveraging continual learning in NLP will become more and more important. As a survey of the existing literature, we could not see that anyone will be made disadvantaged by this work. All the surveyed papers are peer-reviewed and are publicly available. We do not see any risk in this work.

\section{Limitation}
While we tried to cover as many systems as possible, the covered systems are by no means exhaustive. Our main effort has been put on searching and reviewing papers published in main NLP, machine learning and computer vision conferences. However, since continual learning has been growing rapidly, many related papers might have been missed. In our future work, we plan to continually update the paper list and introduce the most recent topics/papers on continual learning to the NLP community in a dedicated website.

\bibliographystyle{named}
\bibliography{ijcai22}

\appendix
\null\newpage 

\appendix
\null\newpage 

\section{NLP Preliminaries}
\label{sec.preliminary}

Almost all NLP systems nowadays are based on Transformer~\cite{liu2019roberta,DBLP:conf/acl/LewisLGGMLSZ20,DBLP:conf/naacl/DevlinCLT19}. This leads to CL techniques specifically for Transformer. Before discussing the detailed CL approaches, it is necessary to discuss these Transformer based NLP techniques.

\textbf{Adapter, prefix, and prompt tuning (light-weight fine-tuning).}  
These approaches (a.k.a., light-weight tuning) add a small number of parameters to a pre-trained Transformer language model (LM). Research has shown that only fine-tuning the added parameters (with the LM frozen) can already achieve similar performance to fine-tuning the whole LM. 
The popular light-weight fine-tuning in CL for NLP include \textit{adapter-tuning}~\cite{Houlsby2019Parameter}, which inserts a fully-connected network to each Transformer layer;
\textit{prefix-tuning}~\cite{DBLP:conf/acl/LiL20}, which prepends some (a hyper-parameter) tunable prefix vectors to the keys and values of the multi-head attention at every layer;
and \textit{prompt-tuning}~\cite{DBLP:conf/emnlp/LesterAC21} (a.k.a., soft-prompt tuning), which adds a sequence of trianable prompt tokens to the end of the original input sequence. In CL, these light-weight tuning methods make the parameter-isolation based methods (Sec.~\ref{sec.parameter_isolate}) more efficient because only a tiny number of parameters are needed to be saved to prevent CF.

\textbf{Instruction (hard prompt).} Hard prompt or instruction is a short piece of text describing the core concept of the task.  Some examples are ``listing nouns'', ``output the nth word or char''~\cite{DBLP:journals/corr/abs-2010-11982} (more can be seen in \cite{DBLP:conf/acl/MishraKBH22}). By using instruction, all tasks can be formatted as a response to a natural language input and thus a generation task. In CL, this helps convert all tasks in the same generation format so that an LM can be trained as both a classifier and a sample generator. Different task-specific instructions can be applied for different tasks to prevent CF. Some datasets may also provide similar/positive instruction that can be used for knowlege transfer (KT).

\section{Details of Systems for CF Prevention}
\label{sec.sys_details_cf}

\subsection{Regularization-based Methods}

\textbf{(1). Regularizing the loss based on parameter importance.}

EWC~\cite{Kirkpatrick2017overcoming} uses the gradient of previous tasks as the importance of parameters for the previous tasks, denoted as $F^{\mathcal{T}}_k$ (i.e., fisher information matrix, where $k$ is the index of parameters). 
$F^{\mathcal{T}}_k$ is then multiplied with the difference between the old parameters and new parameters. {\color{black}The resulting regularization term is added to the loss to penalize the changes made to  important parameters.} Several CL systems in NLP are derived from EWC or leverage the idea to penalize changes to important parameters.

\textbf{DIL (w/o ID)}. ``w/o ID'' means without task-ID.~RMR-DSE \cite{li-etal-2022-overcoming} applies EWC to seq2seq modeling. It converts $F^{\mathcal{T}}_k$ to tunable hyper-parameters so that no additional backward pass is needed after training each task.
AEWC~\cite{DBLP:journals/corr/abs-1712-09943} employs an improved EWC algorithm~\cite{zenke2017continual}, where the importance can be computed in an online manner. The main idea is to replace the fisher matrix with an importance estimator so that the additional re-training and point-estimation can be removed. 


\textbf{CIL.} SRC~\cite{DBLP:conf/naacl/LiuUS19} continually updates the sentence encoder with regularization. The penalty is computed via matrix conceptors which capture the corpus-specific features of each corpus. 

\textbf{(2). Regularizing the loss based on distillation.}

\textbf{CIL.} ExtendNER~\cite{DBLP:conf/aaai/MonaikulCFR21}, CFID~\cite{DBLP:conf/coling/LiZCGZZ22}, CID~\cite{DBLP:journals/corr/abs-2108-04445} and PAGeR~\cite{DBLP:conf/naacl/VarshneyPKVS22} apply \textit{KL-divergence} between the response of the previous model and the current model to achieve distillation. 

\textbf{DIL (w/o ID).} LFPT5~\cite{DBLP:conf/iclr/QinJ22} applies KL-divergence between the saved old prompt and the new prompt. 


\textbf{Other regularization-based methods in CV.}

For example, in learning a new task, OWM~\cite{zeng2019continuous} projects the gradient in the orthogonal direction to the input of the old tasks. \cite{Wang_2021_CVPR} maps the gradient to the null space of the previous tasks to mitigate CF. 
Some others regularize the learning rate to slow down parameter updates or control the geometry of the local minima~\cite{DBLP:conf/iclr/EbrahimiEDR20,DBLP:conf/nips/MirzadehFPG20}.



\subsection{Replay-based Methods}

\textbf{(1). Replaying raw samples in training.}

\textbf{CIL.} CFID~\cite{DBLP:conf/coling/LiZCGZZ22} replays raw samples with a dynamic weighting to control the sample weight. CID~\cite{DBLP:journals/corr/abs-2108-04445} further considers the imbalance between the rich new data and the small old data. CID also leverages the idea in the imbalanced learning community (Inter-class margin loss and cosine normalization) to alleviate the data imbalance issue in replay.

\textbf{DIL (w/o ID; DP).} ``DP'' refers to domain-adaptive pre-training.
Continual-T0~\cite{scialom2022continual}
saves 1\% of the previous task samples in the memory buffer. It can preserve previous knowledge well and even preserve the zero-shot performance of the T0 model. ELLE~\cite{qin2022elle} uses a large memory buffer (around 1G per task).

\textbf{CIL.} IDBR~\cite{huang2021continual}  proposes a memory selection rule based on K-means to store as fewer samples as possible. That is, it
saves only examples closest to each cluster centroid.

\textbf{(2). Replaying raw samples in training and inference.}

\textbf{CIL (Online) and DIL (w/o ID; online).}
MBPA++ \cite{DBLP:conf/nips/dAutumeRKY19} and Meta-MBPA++ \cite{wang2020efficient} selectively use the saved old samples in inference. During inference, the representation of a test sample is used to retrieve $K$ nearest saved samples in the replay memory. Based on the $K$ samples, gradient updates are performed to achieve sample-specific fine-tuning. The fine-tuned network is then used to output the final prediction for the test sample. 

\textbf{(3). Replaying prototypes. }

\textbf{TIL. }MeLL~\cite{DBLP:conf/kdd/0001PLCQZHCL021} stores class prototype representations for meta-learning (Sec.~\ref{sec.meta}). Both the stored prototypes and current task feature representations are used to prevent CF.

\textbf{(4). Optimizing on the replay memory.} Details are given in main text.

\textbf{(5). Replaying generated pseudo samples.}

\textbf{TIL.} PCLL~\cite{DBLP:journals/corr/abs-2210-07783} uses prompt and conditional variational autoencoder (CVAE) to generate previous samples. It saves the task-specific prompt for each task so the saved previous task prompt can be used to help condition the LM when generating previous task samples.

\textbf{DIL (w/ ID).} {``w/ ID'' refers to with task-ID.}
LAMOL \cite{sun2020lamol} and ACM~\cite{zhang2022continual} convert all tasks into generation tasks following \cite{McCann2018decaNLP} and train the underlying LM (GPT-2) so that it can do both task learning and sample generation. When a new task comes, LAMOL or ACM first leverages the LM to generate old task samples and then train the ``old'' and the new samples together to avoid CF. Different from LAMOL, ACM makes use of adapters and selectively shares them to further enable KT (Sec.~\ref{sec.appro_kt}). 

\textbf{CIL.} PAGeR~\cite{DBLP:conf/naacl/VarshneyPKVS22} replays the generated samples by using instructions (hard prompts) containing both current and previous intent words.

\textbf{DIL.} Similarly, LFPT5~\cite{DBLP:conf/iclr/QinJ22} leverages prompt and T5 to achieve both task learning and sample generation. Recall that LFPT5 also has the prompt distillation (Sec.~\ref{sec.reg_cf}) to prevent forgetting, so LFPT5 is a combination of regularization-based and replay-based method. 

\subsection{Parameter-isolation Based Methods}
\label{sec.paramter_isolate}

\textbf{(1). Fixed architecture.}

\textbf{(i). Task masking.} 

\textbf{HAT} initializes a task embedding $\bm{e}^{(t)}_l$ for each task $t$ and each layer $l$ in the network. A $\texttt{sigmoid}$ function is used as a pseudo-gate/step function along with a large positive number $s$ (hyper-parameter). A mask $\bm{m}_l^{(t)}$ is given by $
\bm{m}_l^{(t)} =  \sigma (s\bm{e}^{(t)}_l)
$.
During training, the mask is  element-wise multiplied with the output of each layer $l$. 
In backward propagation, HAT blocks the used neurons (indicated by the mask) by previous tasks via multiplication of the inverse of the mask with the gradient to avoid CF.  

There are a couple NLP systems adopt this idea, including,

\textbf{TIL.} CTR~\cite{ke2021achieving} and B-CL~\cite{ke2021adapting}

\textbf{DIL (w/o ID; DP).}  CPT~\cite{ke2022cpt}. 

\textbf{DIL (w/o ID).} CLASSIC~\cite{ke2021Classic}

\textbf{(ii). Parameter generation.} 

Hypernet~\cite{von2019continual} builds an generated network $g$, which takes the task representation $z$ as input to generate the model parameters for a task solver network $f$. Since the generated network $g$ itself is exposed to CF, Hypernet further imposes a regularization (it is thus also a regularization-based based method) to penalize the changes to the weights of $g$. 

\textbf{TIL. }CLIF~\cite{DBLP:conf/emnlp/JinLR021} adopts this idea and uses Hypernet to generate parameters for the adapters.

\textbf{(iii). Subnetwork masking.}

SupSup~\cite{wortsman2020supermasks} trains a binary mask network via a ``straight-through'' trick for each task so that different tasks are totally isolated from each other. However, its subnetwork or mask potentially has the same size as the original model. Extra training and saving tricks are needed to make it efficient.

\textbf{(2). Dynamic architecture.}

\textbf{DIL (w/ ID).}
C-PT~\cite{zhu2022continual} adds a prompt for each task. TPEM~\cite{DBLP:conf/acl/GengYXSX020} expands the network whenever a new task comes.

\textbf{DIL (w/o ID).} AdapterCL~\cite{madotto2020continual} adds one set of adapters for each task. It further infers the task-id in testing via perplexity.

\textbf{DIL (w/o ID; DP).} DEMIX~\cite{gururangan2021demix} adds adapters when a new task comes. In testing, the prediction of a test instance is given by the weighted sum (according to perplexity) of all pre-trained adapters. ELLE~\cite{qin2022elle} expands layers whenever a new task comes. It uses a large replay data (Sec.~\ref{sec.replay}) to ensure that the LM can deal with all previous and current tasks.  

\textbf{CIL.} ProgM~\cite{shen-etal-2019-progressive} expands the network for each new task and also transfers all the learned knowledge to the expanded component so that 
it can directly use only the last expanded component in testing.

\textbf{(3). Parameter pool.}

L2P~\cite{wang2021learning} initializes a pool of prompts and selects different prompts for different tasks based on cosine similarity.

\subsection{Instruction-CF Based Methods}

\textbf{DIL (w/ ID).} ConTinTin~\cite{DBLP:conf/acl/0001LX22} uses different instructions for different tasks to condition the LM.
ENTAILMENT~\cite{DBLP:conf/naacl/XiaYFY21} inputs both the sentence and intention label words as instruction and converts intent classification to binary classification, which helps alleviate forgetting.


\section{Details of Systems for Knowledge Transfer}
\label{sec.sys_details_kt}

\subsection{Similarity-based Methods}

\textbf{(1). Importance-proxy.} 

\textbf{DIL (w/ ID).} ACM~\cite{zhang2022continual}, which learns an adapter for each task and trains a mixing coefficient to detect which adapter can be reused by a new task. 
ACM prevents CF via a pseudo-replay method.

\textbf{(2). Feature-proxy.}

\textbf{(a). Capsule Network (CapsNet).}

\textbf{TIL.} CTR~\cite{ke2021achieving} and B-CL~\cite{ke2021adapting} are based on this idea. We know that they both use a parameter-isolation method to avoid CF. They further compare the task similarity via CapsNet. In B-CL and CTR, each task uses a capsule in the lower level (called ``task capsule''). The system leverages the routing algorithm to group similar tasks and their shareable features: if two tasks are similar, the capsules corresponding to the tasks are allowed to update (in CTR, there is a binary gate. In B-CL, a weight based on similarity is employed to control how much can be updated) so that knowledge transfer (KT) can be achieved. 
Note that the parameter-isolation based method like HAT (Sec.~\ref{sec.parameter_isolate}) has parameter sharing (but fixed) across tasks, this may cause CF for dissimilar tasks that share parameters with those similar tasks.

\textbf{(b). Contrastive learning.}

\textbf{DIL (w/o ID). }~CLASSIC~\cite{ke2021achieving} uses this idea. Again, CLASSIC prevents forgetting via a parameter-isolation method (Sec.~\ref{sec.parameter_isolate}). It encourages KT via contrastive learning. To this end, CLASSIC generates an augmented view from all previous tasks via self-attention and uses contrastive loss to help learn the shared knowledge. 

\subsection{Replay-based Methods}
\label{sec.meta}

\textbf{(1) Regularization-based}

\textbf{CIL.}  IDBR~\cite{huang2021continual}. We have seen that IDBR leverages a replay memory to prevent CF (Sec.~\ref{sec.replay}). To encourage the LM to learn task general knowledge, it further regularizes the loss by adding a next sentence prediction task.

\textbf{(2) Meta-learning-based Methods}

In meta-learning, there are two stages. One is \textit{meta-training} stage where the goal is to learn good representation that can quickly adapt to new tasks. The other one is \textit{meta-testing} stage where the goal is fine-tuned (usually few-shot) the meta-trained model to do well on the target tasks. In meta-training stage, a set of source tasks is defined and each task has its training and validation data, called \textit{support} and \textit{query} set correspondingly. Note that support and query set have their own training and testing set because its goal is learn general representation for the meta-testing. In meta-testing stage, there is another set of tasks which are the target tasks. 

By defining the source task as the new task data and the query tasks as all data in the old and new tasks, forgetting prevention and knowledge transfer are naturally embedded in the loss of meta-learning~\cite{DBLP:conf/iclr/RiemerCALRTT19,DBLP:journals/corr/abs-2007-13904}. Since the query sets need to contain the old tasks data, meta-learning-based methods are always replay-based methods.

Specifically, meta-learning can be viewed as a bi-level optimization problem as follows: 
\begin{equation} 
\begin{split}
\label{eq:meta}
& w^* = \text{argmin}_w\sum_{i=1}^M\mathcal{L}^{\text{meta}}(\theta^{*(i)}(w),w,\mathcal{D}_{\text{query}}^{(i)}) \\
& \text{s.t. } \theta^{*(i)}(w) = \text{argmin}_\theta \mathcal{L}^{\text{task}}(\theta,w,\mathcal{D}_{\text{support}}^{(i)})
\end{split}
\end{equation}
where $\theta$ is the network parameter and $w$ is the learning strategy of $\theta$. While $w$ can be anything that affect $\theta$ (e.g., the learning rate), a typical $w$ is the initialized parameters. $w$ is the key in meta-learning because it specifies ``how to learn'' $\theta$ and it is called \textit{meta-representation}.

While there is ``task'' in CL and meta-learning, they are very different. To distinct them, we call the task in meta-learning the meta-task and the task in CL the CL-task.  In continual learning, the set of meta-tasks are usually defined as the data in each batch. For example, current CL-task data in each batch is used as support sets while the combination of current CL-task data and memory data in each batch is used as query sets. One exception is the online CL methods where usually each sample is used as a meta-task. 

This idea has been used in several NLP systems.

\textbf{TIL.} MeLL~\cite{DBLP:conf/kdd/0001PLCQZHCL021} has a memory network ($\theta$) and an LM ($w$). The memory contains all the learned class prototype representations. Its output is concatenated with the output of the LM and fed into a fusion network to perform classification. When training the support set, the current CL-task's class prototypes in the memory network are updated while LM is fixed. When training on the query set, the LM is fine-tuned while network is fixed. 
As a result, the LM learns the general/shared information (for KT) and the memory network learns the task specific information (for dealing with CF). 

\textbf{DIL (w/o ID; Online) and CIL (Online).} Meta-MBPA++~\cite{wang2020efficient} uses an additional network as  $\theta$ in the inner loop to retrieve samples from the replay memory. An LM as text encoder is used as $w$ to generalize well across all tasks in the outer loop.

\textbf{(3) Instruction-based Methods}

\textbf{DIL (w/ ID).} C-PT~\cite{zhu2022continual} fuses the previous instruction (the ``query'' in a state tracking task) to enable the model to learn a general skill. 

\textbf{TIL.}
ConTinTin~\cite{DBLP:conf/acl/0001LX22} uses a task-specific instruction to prevent CF (Sec.~\ref{sec.instruct_cf}), and further uses the dataset provided positive instructions/samples to enable knowledge transfer (KT).

\section{Details of Systems for Inter-task Class Separation}
\label{sec.sys_details_tsf}

Replay-based methods~\cite{huang2021continual,DBLP:conf/aaai/MonaikulCFR21,DBLP:conf/coling/LiZCGZZ22,DBLP:journals/corr/abs-2108-04445,DBLP:conf/naacl/VarshneyPKVS22,DBLP:conf/iclr/QinJ22,wang2020efficient,DBLP:conf/nips/dAutumeRKY19,scialom2022continual,qin2022elle} can help deal with ICS to some extent because it allows the model to see a limited number of past data. Below are some post-prediction methods used in DIL systems. 

\textbf{DIL (w/o ID).} AdapterCL~\cite{madotto2020continual} compares the perplexity of each task model (they have previous task models because they are parameter-isolation based methods) using the test samples. The lowest perplexity model will be used to conduct the inference. RMR-DSE~\cite{li-etal-2022-overcoming} compare the feature similarity between test samples and a set of domain-shifting vectors (representative vector extracted from the difference between past model representation and current model representation).

\textbf{DIL (w/o ID; DP).} DEMIX~\cite{gururangan2021demix} uses perplexity to predict task-ID, similar to AdapterCL.

\section{Continual Learning Evaluation}
\label{sec.evaluate}
CL evaluation mainly assesses (1) the average performance of all learned tasks, (2) the rate of forgetting 
and (3) the effect of knowledge transfer. 
Below, we present some reference baselines and evaluation metrics. 

\subsection{Reference Baselines}

Several reference (or control) baselines are commonly used as the lower or upper bounds for CL. 

\textbf{Naive continual learning (NCL).} This baseline refers to a system that performs CL without any mechanism to deal with forgetting~\cite{ke2021achieving,ke2021adapting,DBLP:conf/iclr/QinJ22}. All parameters can be freely updated in learning the new task. This baseline is expected to have the worst CF and {\color{black}the base performance for KT}, and thus is often regarded as the lower bound of a CL system. 

\textbf{Train a separate model for each task (ONE).} Opposite to NCL, ONE trains each task separately as an independent model~\cite{ke2021achieving,ke2021adapting,DBLP:conf/kdd/0001PLCQZHCL021}. It is thus not a CL setting and has no CF or KT. It is usually regarded as the control to assess whether a CL system has KT or CF.

\textbf{Multi-task learning (MTL).} MTL is another non-CL reference baseline that is widely regarded as the upper bound of TIL. It requires a multi-head setup.  

\textbf{Aggregate.} For those systems using a single-head setup or CIL, this reference baseline learns the classes of all tasks as a single learning task. This is regarded as the upper bound for DIL, and CIL. 

\subsection{Evaluation Metrics}

\textbf{Average Accuracy}~\cite{chaudhry2018riemannian} ($A_{\mathcal{T}}$) measures the performance of a CL method after all $\mathcal{T}$ tasks have been learned, i.e.,
$
    A_{\mathcal{T}} = \frac{1}{\mathcal{T}}\sum^{\mathcal{T}}_{t=1}a_{\mathcal{T},t}
$,
where $a_{\mathcal{T},t}$ refers to the performance of the model on the testing set of task $t$ after the model has continually trained all $\mathcal{T}$ tasks.

\textbf{Average Incremental Accuracy}~\cite{Rebuffi2017,Lopez2017gradient} is a metric in CIL and derived from average accuracy. It is simply the average over the average accuracy of each task (a set of classes in CIL). One can also choose to curve the average accuracy of each task in a figure rather than giving a single number of average incremental accuracy.


\textbf{Forgetting Rate}~\cite{chaudhry2018riemannian} ($F_{\mathcal{T}}$) measures how much knowledge has been forgotten across the first $\mathcal{T}-1$ tasks, i.e.,
$
    F_{\mathcal{T}} = \frac{1}{\mathcal{T}-1}\sum^{\mathcal{T}-1}_{t=1}(a_{t,t}-a_{\mathcal{T},t})
$,
where $a_{t,t}$ is the test accuracy task $t$ right after it is learned, and $a_{\mathcal{T},t}$ is the accuracy of task $t$ after training the last task $\mathcal{T}$. We average over all trained tasks except the last one as the last task has no forgetting. The higher the forgetting rate is, the more
forgetting it has. Negative rates indicate positive knowledge transfer.

\textbf{Forward Transfer}~\cite{ke2020mixed} ($\text{FWT}_{\mathcal{T}}$) measures how much forward transfer has happened to a new task after learning the task, i.e.,
$
    \text{FWT}_{\mathcal{T}} = \frac{1}{\mathcal{T}}\sum^{\mathcal{T}}_{t=1}(a_{t,t}-a_{0,t})
$,
where $a_{0,t}$ refers to the performance of training task $t$ individually (i.e., the accuracy of ONE for the task). Note that there is another way to measure forward transfer. It measures whether a learned network contains some useful knowledge for the new task~\cite{Lopez2017gradient}, i.e., 
$
    \text{FWT}'_{\mathcal{T}} = \frac{1}{\mathcal{T}-1}\sum^{\mathcal{T}}_{t=2}(a_{t-1,t}-r_t)
$,
where $r_t$ refers to the accuracy of task $t$ using a randomly initialized network. {\color{black}Unlike $\text{FWT}_{\mathcal{T}}$, $\text{FWT}'_{\mathcal{T}}$} does not tell how much forward transfer actually happens after learning the new task.

\textbf{Backward Transfer} ($\text{BWT}_{\mathcal{T}}$) measures how much the continual learning of subsequent tasks influences the performance of a {learned} task, i.e.,
$
    \text{BWT}_{\mathcal{T}} = \frac{1}{\mathcal{T}-1}\sum^{\mathcal{T}-1}_{t=1}(a_{\mathcal{T},t}-a_{t,t})
$.
Positive $\text{BWT}_{\mathcal{T}}$ indicates the subsequent task learning can improve the performance of previous tasks.

\end{document}